\documentclass[letterpaper, 10 pt, conference]{ieeeconf} 

\IEEEoverridecommandlockouts                   
\usepackage{graphicx}
\usepackage{caption}
\usepackage{subcaption}
\usepackage{url}
\usepackage{xcolor}

\usepackage{amssymb}
\usepackage{booktabs}
\usepackage{tabularx}
\usepackage{tikz}
\usepackage{microtype} 
\usepackage{pifont}
\usepackage{xcolor}
\usepackage{booktabs} 
\usepackage{siunitx} 
\usepackage{multirow}

\title{\LARGE \bf
Stonefish: Supporting Machine Learning Research in Marine Robotics
}

\author{Michele Grimaldi$^{1*}$,  Patryk Cie\'{s}lak$^{2*}$, Eduardo Ochoa$^{2}$, Vibhav Bharti$^{1}$, Hayat Rajani$^{2}$, Ignacio Carlucho$^{1}$, \\Maria Koskinopoulou$^{1}$, Yvan R. Petillot$^{1}$ and Nuno Gracias$^{2}$
\thanks{*Both authors have contributed equally to authoring this work.}
\thanks{$^{1}$Oceans Systems Lab (OSL), Heriot-Watt University, EH144AS, Edinburgh, UK}
\thanks{$^{2}$Underwater Vision and Robotics Lab (CIRS), University of Girona, 17003, Girona, Spain}
}

\begin{document}
\maketitle
\thispagestyle{empty}
\pagestyle{empty}

\begin{abstract}
Simulations are highly valuable in marine robotics, offering a cost-effective and controlled environment for testing in the challenging conditions of underwater and surface operations. Given the high costs and logistical difficulties of real-world trials, simulators capable of capturing the operational conditions of subsea environments  have become key in developing and refining algorithms for remotely-operated and autonomous underwater vehicles.
This paper highlights recent enhancements to the Stonefish simulator, an advanced open-source platform supporting development and testing of marine robotics solutions. Key updates include a suite of additional sensors, such as an event-based camera, a thermal camera, and an optical flow camera, as well as, visual light communication, support for tethered operations, improved thruster modelling, more flexible hydrodynamics, and enhanced sonar accuracy. These developments and an automated annotation tool significantly bolster Stonefish's role in marine robotics research, especially in the field of machine learning, where training data with a known ground truth is hard or impossible to collect. 
\url{https://github.com/patrykcieslak/stonefish}
\end{abstract}

\section{Introduction}

In the rapidly advancing field of robotics, simulations play an important role in research and practical applications, offering a controlled and cost-effective environment for testing and development~\cite{liu2021role}. This is particularly true in the realm of marine robotics, where the harsh and unpredictable conditions of underwater environments present significant challenges ~\cite{Grimaldi2023InvestigationOT}. Remotely-operated vehicles (ROVs) and autonomous underwater vehicles (AUVs) are integral to a wide range of applications including oceanographic research, underwater infrastructure inspection, and deep-sea exploration~\cite{petillot2019underwater,10801590}. The development and deployment of subsea robots demand rigorous testing to ensure reliability and performance under various environmental conditions. However, real-world underwater testing is often impractical due to high costs, logistical complexities, and the potential for damage to expensive equipment. Consequently, simulators capable of capturing the features of underwater environments have emerged as indispensable tools in the field of subsea robotics \cite{sim_review}\cite{adetunji2024digitaltwinssurfaceenhancing}. Moreover, new algorithms based on deep learning and other machine learning approaches are being extensively used nowadays and their development is hindered by the lack of sufficient amounts of annotated data. 
This data includes images from cameras operating in and out of the visual spectrum, like the thermal camera, from event-based sensors, and from sonars, as well as, e.g., navigation ground truth.

\begin{figure}[t]
    \centering
    \includegraphics[width=0.95\linewidth, height=7.5cm,  trim=0cm 7.5cm 0cm 7.5cm,clip]{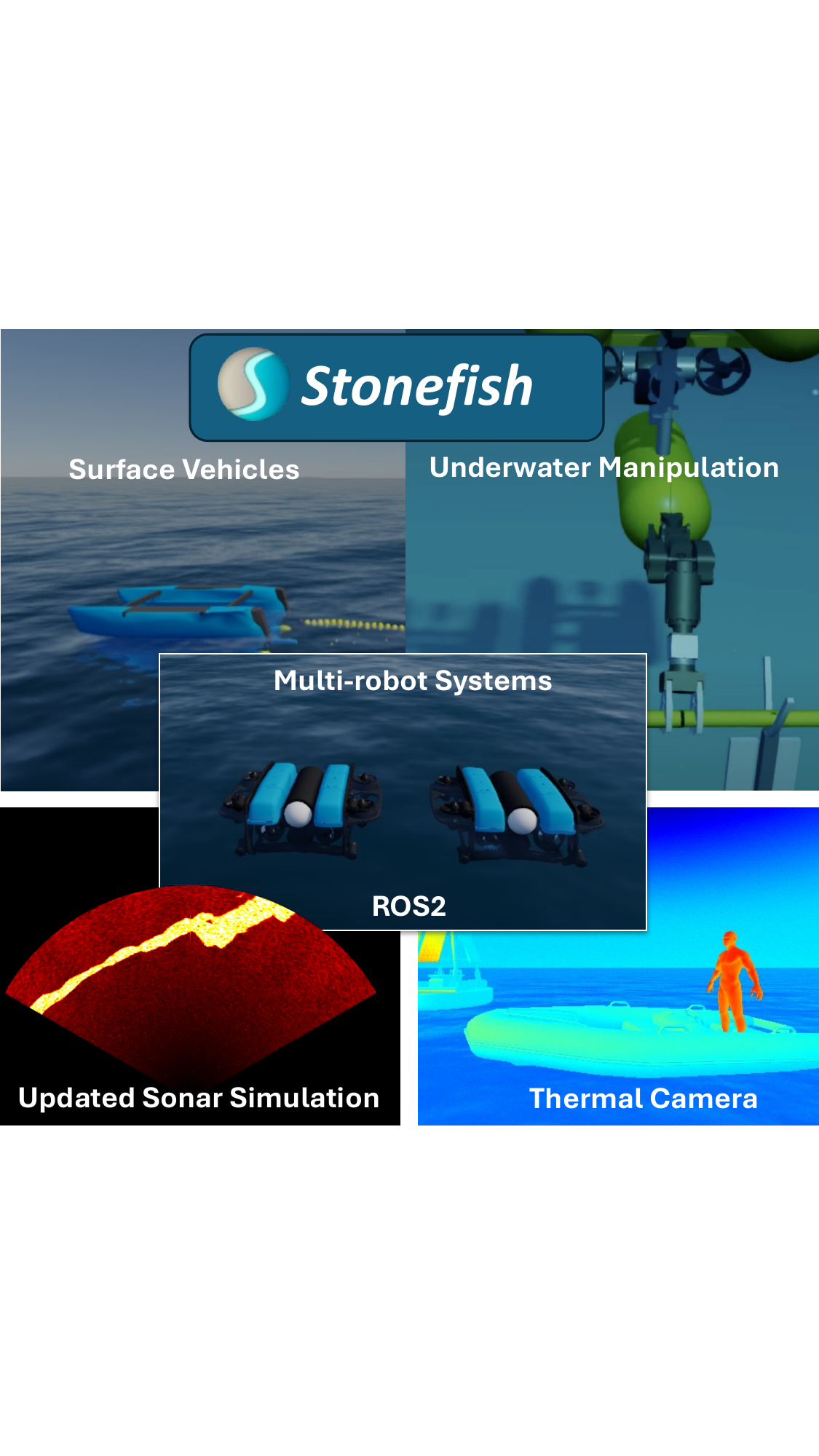}
    \caption{Stonefish: key features and improvements.}
    \label{fig:stonefish}
\end{figure}

This paper presents recent advancements in the Stonefish simulator~\cite{cieslak2019stonefish}, a state-of-the-art marine robotics simulation platform. Significant improvements, as shown in Fig.~\ref{fig:stonefish}, span across various key areas, making it one of the most versatile and accurate simulators for marine and especially underwater robotics. Notable enhancements include the integration of new GPU-accelerated sensors, implementation of visual light communication (VLC) and enhancement of sonar simulation. The aforementioned new sensors include an event-based camera (EBC), a thermal camera and an optical flow sensor. The existing forward-looking sonar (FLS) has been significantly upgraded, resulting in increased accuracy and reliability of the simulated data. 
All these additions and improvements notably expand the platform’s capabilities in sensor simulation, thus opening new possibilities for generating datasets for machine learning research. Additionally, the simulator now supports linking robots with tethers, providing more realistic behaviour of robots involved in subsea operations. The introduction of configurable thruster dynamics has extended the flexibility and control over the utilised mathematical models, enhancing insights into vehicle dynamics and control strategies. Furthermore, the simulator now includes an automatic annotation tool, streamlining data analysis and improving the efficiency of training and testing environments. These comprehensive upgrades solidify the position of Stonefish as a leading tool for marine robotics research and development.

\begin{figure*}[tp] 
\centering
    \includegraphics[width=\textwidth]{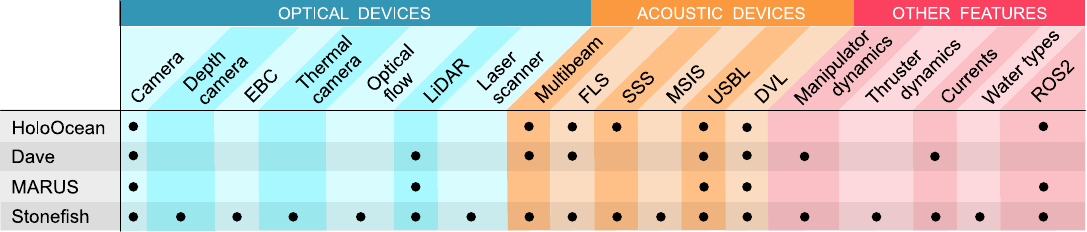}
    \caption{Comparison of availability of the most important features in selected simulators.}
    \label{fig:comparison}
\end{figure*}

\section{Related Work}

\subsection{Classical simulators}

One of the most well-known underwater robotics simulators is the UWSim \cite{dhurandher2008uwsim}.  It enabled the simulation of underwater inspection and intervention missions with one or more robots. One of its main advantages is that it delivers a ROS (Robot Operating System \cite{quigley2009ros}) interface allowing for easy integration with ROS-based software architectures, commonly used in robots developed by the scientific community. 
However, in practice, UWSim is mostly used as a visualisation tool and for simulating underwater sensors. It lacks accurate simulation of dynamics and hydrodynamics of vehicles and it does not support simulation of manipulator dynamics (only kinematics). Moreover, this simulation tool has been deprecated and not updated in several years.

A more modern approach is to use the ROS standard simulator Gazebo \cite{gazebo}. Gazebo relies on DART and Bullet Physics for the physics and OGREv2 for the graphics. It lacks native support for hydrodynamics and has poor rendering quality. It can be used either with a custom dynamics code or combined, e.g., with the UUV Simulator plugin \cite{Manhaes_2016}, which adds simple hydrodynamics to the Gazebo physics engine. However, this solution is lacking the simulation of manipulator hydrodynamics and buoyancy. 

Another simulator, Project DAVE \cite{zhang_dave_2022} was built upon the foundation of UUV Simulator \cite{manhaesUUVSimulatorGazebobased2016} and \texttt{ds\_sim} tool package \cite{WhoidslDs_simBitbucket}, featuring a comprehensive array of environments, robots, sensors, and demonstrations, with a particular focus on underwater manipulation tasks.

Another classical simulator is the Marine Systems Simulator~\cite{perez2006overview}. It is a Matlab/Simulink toolbox that delivers accurate dynamic simulation of underwater vehicles. However, the lack of visualisation and no direct interface with ROS makes it unpopular in the community.

\subsection{Unity-Based Underwater Simulators}
Several underwater simulators have been developed using the Unity3D game engine, taking advantage of its realistic visuals, accurate rigid-body physics, and user-friendly development environment. Notable examples include:

\begin{itemize}

\item UWRoboticsSimulator~\cite{UwRoboticsSim}: Specifically developed for marine environments, this simulator supports both ROS and ROS2 and is still actively maintained. Despite its active development, it lacks essential sensors like LiDAR, sonar, and DVL.

\item URSim~\cite{URSim}: This simulator also focuses on the marine environment but the last supported ROS version was Kinetic, and development has ceased since 2019. URSim is therefore outdated and lacks support for recent ROS versions.

\item MARUS~\cite{Marus}: Developed to address the limitations of existing Unity-based simulators, it leverages the engine's capabilities to offer high-end graphics, computation parallelized on GPU, simplicity of test scene design, and many out-of-the-box tools for physics simulation. Unity’s popularity is supported by a broad and versatile community that provides numerous examples and solutions for common problems in game design. MARUS extends Unity3D’s physics engine with water body physics (buoyancy, waves, etc.), simulation of various sensors, propulsion emulation, and visualization tools.

\end{itemize}

Unity3D's strengths in realistic visuals and ease of development have made it a popular choice for simulator development. However, as an engine targeted at game development, it delivers many tools that are not necessary for researchers and were created for artists, who do not have the coding experience. The amount of additional tools results in vast download sizes. Moreover, many of the advanced features of Unity3D are not free but require a commercial license.

\subsection{Unreal Engine-Based Underwater Simulators}

Unreal Engine (UE) is probably the most powerful game engine available today. It is known for its high-fidelity graphics and GPU-accelerated physics engine. Key simulators include:

\begin{itemize}
\item HoloOcean~\cite{potokar2022holoocean}: Built on top of UE4, HoloOcean provides high-fidelity imagery and accurate dynamics using the PhysX physics engine. It features a simple Python interface, making it easy to install and use across various systems. HoloOcean augments Holodeck with accurate underwater dynamics, multi-agent communications, and realistic imaging sonar implementations. 

\item UNav-Sim~\cite{amer2023unav}: The first open-source underwater simulator based on Unreal Engine 5 (UE5), UNav-Sim offers superior rendering quality essential for developing AI and vision-based navigation algorithms for underwater vehicles. It supports robotics tools such as ROS2 and autopilot firmware, making it suitable for robotics research and development. UNav-Sim uses open-source AirSim \cite{AirSim} extensions to add custom vehicle models and integrates AirSim with UE5. This simulator excels in creating photorealistic environments and supports a wide range of underwater scenarios and models, making it highly effective for developing vision-based localization and navigation methods for underwater robots.
\end{itemize}

The use of UE for simulators brings high-quality visuals and highly parallelised physics calculations to underwater simulation. Moreover, UE can be used completely free of charge.
However, similarly to the Unity3D, the UE requires a download of a vast amount of tools that are targeted at artists and the development of a custom simulator based on UE requires deep understanding of its code-base and thus a lot of work hours. Moreover, the default physics engine (PhysX) is non-deterministic, resulting in different simulation results in each run.
In both cases, the Unity3D and the UE, the researchers must rely on the implemented physics engines or replace them with their own implementations designed specifically for underwater environments. All the sensor simulations have to be implemented as well, e.g., in form of plugins.

\subsection{Stonefish}
Stonefish is an open-source C++ library combining a physics engine, based on Bullet Physics, and a custom lightweight rendering pipeline utlilising the OpenGL API. It is directed towards researchers in the field of marine robotics but can also be used as a general-purpose robot simulator.

The physics engine marries the functionality of the Bullet's collision detection and multi-body dynamics based on Featherstone's algorithm \cite{featherstone}, with custom treatment of material interaction, buoyancy and fluid dynamics. The aforementioned interaction between materials is defined by a table of static and dynamic friction coefficients, which contrast the unrealistic single material coefficients found in other simulators. Its advanced hydrodynamic computations are based on the actual geometry of bodies, to allow for effects not possible when directly utilising the classical rigid-body models described by symbolic quantities. One of the situations where this approach delivers notably more natural behaviour is when local currents are defined in the ocean, that act on the robot while it is passing through, and cause it to rotate when entering the current, instead of moving sideways, as the water velocity is sampled at each triangle of the mesh. 

The rendering pipeline, developed from the ground up, delivers a realistic rendering of the atmosphere, ocean and underwater environment. This approach was chosen to minimise external dependencies and implement a solution tailored to the needs of the simulator.

In this paper, we present a recent version of Stonefish, which bridges the gap between simulation and real-world performance. Since the first publication~\cite{cieslak2019stonefish} the simulator has significantly evolved and is being constantly developed, based on the needs of the growing community of users. This new version introduces several enhancements and new features, mentioned briefly in the introduction, which are described in the subsequent sections. Formerly delivered with a ROS1 interface now Stonefish also supports interaction with ROS2 middleware via the new stonefish\_ros2 package.

Figure~\ref{fig:comparison} compares the functionality of the main, actively developed marine simulators. Stonefish clearly leads among these platforms, delivering the richest sensor suite and the support for realistic manipulator and thruster dynamics, important in control design research.

\section{Sensors}
\subsection{Imaging Sonar}

Since there is a need to get realistic sonar images to test the algorithms with data close to real-world scenarios, we modified the sonar's shaders to improve the output. 

The process begins with initializing histogram bins for each beam and processing the beam samples, ensuring that the range and intensity data adhere to specified parameters. These parameters include the number of beams in the sonar array, which influences resolution and coverage, as well as range settings such as the minimum and maximum distances from which data is collected and the range step, which is the interval between successive range measurements. Gain is applied to amplify the sonar signal's intensity, simulating varying signal strengths.
The shader algorithm also incorporates time of flight (TOF) to enhance simulation accuracy. This parameter influences the intensity of the sonar return signal, providing information about the distance and physical properties of objects underwater. To account for the natural decrease in signal intensity with distance, the algorithm calculates a distance weight as the inverse square of the range. This weight compensates for the decrease in intensity due to geometric spreading and absorption, ensuring that the histogram reflects the actual reflectivity or strength of the targets, rather than being biased by their distance from the sonar.
Noise and modulation are introduced to enhance realism, with noise coordinates calculated and Perlin noise applied to create realistic variations. Noise seeds and noise standard deviation values are used to determine the spread and intensity of the noise, mimicking real-world variations. The algorithm also accounts for hold factors and ghosting effects. Hold factors determine the influence of previous frames on the current frame, simulating the persistence of sonar echoes. The ghosting factor adjusts the intensity of ghosting effects, simulating the fading and trailing of echoes over time.
Beam pattern noise is included to account for inherent imperfections in the sonar beam, and intensity modulation is adjusted based on the proximity and orientation of detected objects. 

\begin{figure}[t] 
\centering
    \includegraphics[width=.5\linewidth]{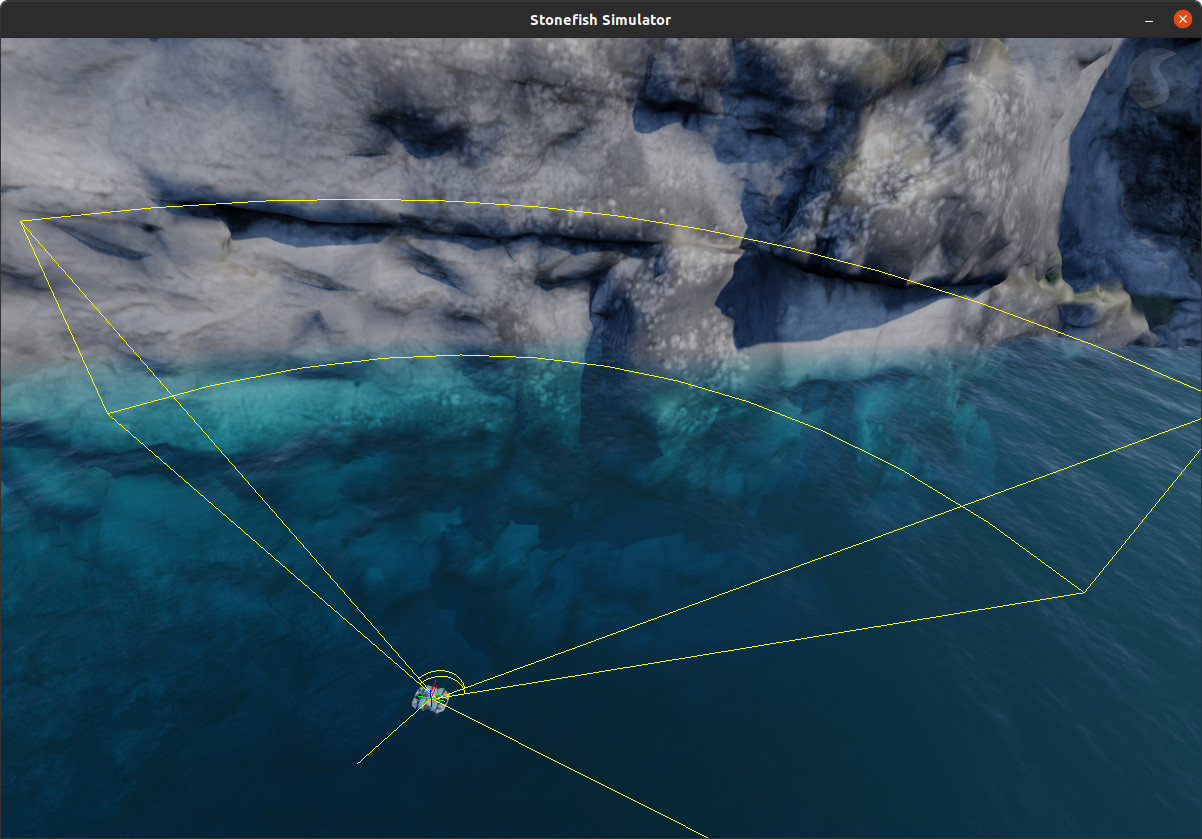}
    
        \vspace{0.1cm}
        
\centering
	\begin{subfigure}[t]{.48\linewidth}
		\includegraphics[width=\textwidth]{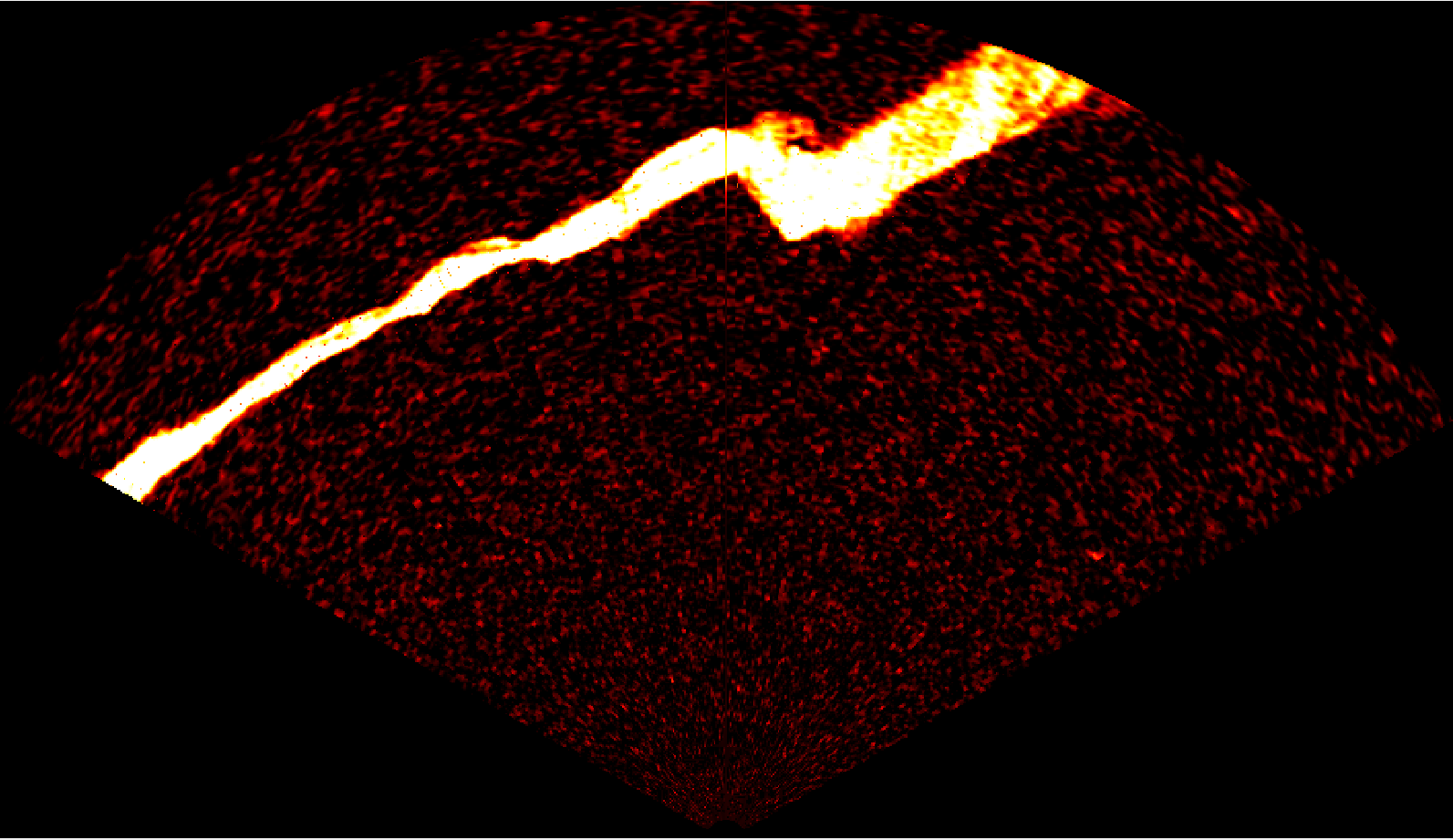}
		\caption{Legacy sonar output}
		\label{fig:1A}
	\end{subfigure}
    \hfill
	\begin{subfigure}[t]{.48\linewidth}
		\includegraphics[width=\textwidth]{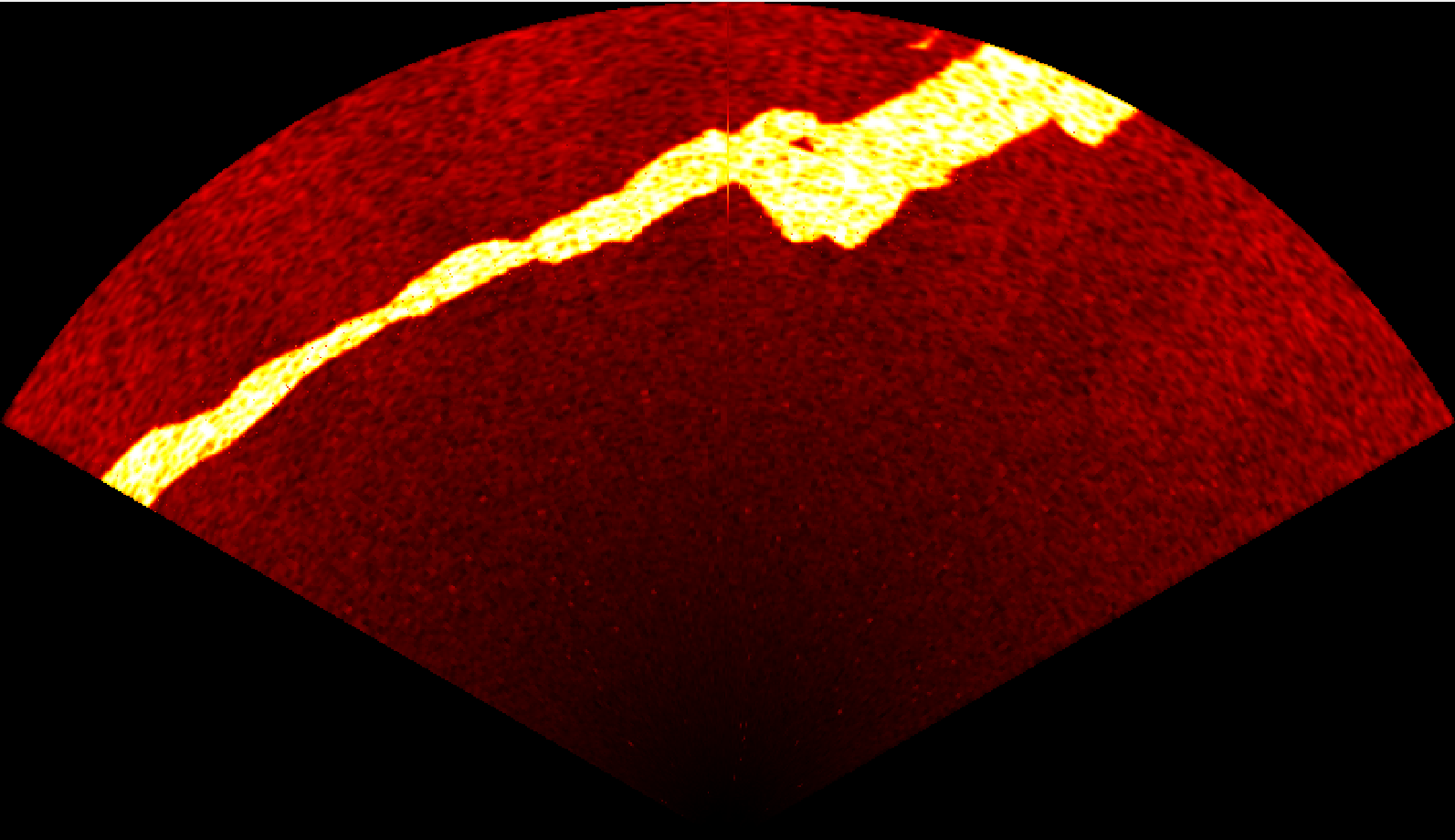} 
		\caption{New sonar output} 
		\label{fig:1B}
	\end{subfigure}
    \caption{Sonar output from a simulated environment presented in the top image, which also shows sonar's field of view.}

\end{figure}

The forward-looking sonar is indeed parametrizable, and in our case, we decided to set the aforementioned coefficients so as to get images closely resembling the output of a Gemini Tritech 1200ik. Figures \ref{fig:1A} and \ref{fig:1B} depict the improvement of the imaging sonar output.

\subsection{Event-Based Camera}
Underwater conditions, characterized by dynamic lighting, scattering, and absorption, make traditional vision sensors less effective. Event-based cameras (EBCs), with their high dynamic range and ability to capture fast changes without motion blur, are particularly suited for these challenging conditions. We implemented the concepts used in the ESIM \cite{Rebecq18corl} EBC simulator in Stonefish, utilising GPU acceleration through OpenGL shaders. The original implementation is fully CPU-based.

The EBC operation is based on monitoring the change of each pixel's luminance independently. Instead of direct luminance measurements, the camera is actually measuring the logarithm of luminance (logL). It is characterised mainly by two quantities: the contrast threshold and the refractory period. The former one is defined separately for negative and positive polarity events and is subject to Gaussian noise. The sensor generates events when two conditions are met: the change of the measured logL is greater than the contrast threshold and the time from the last event is greater than the refractory period. The events are asynchronous and the camera is lacking a typical notion of a framerate.

\begin{figure}[t]
	\centering
	\includegraphics[width=\linewidth]{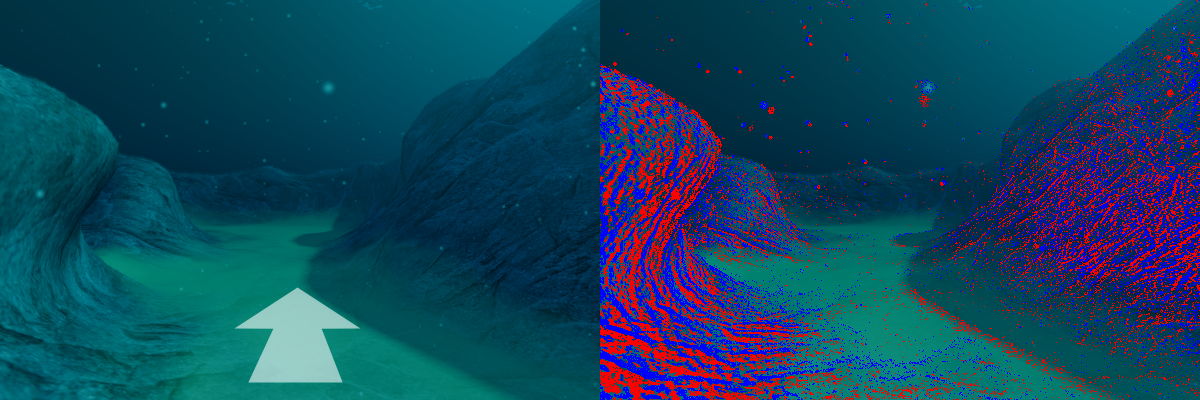}
	\caption{On the left a view from a colour camera (arrow symbolises direction of camera movement); On the right the corresponding EBC events based on the sensor's motion.} 
		\label{fig:evB}
\end{figure}

Implementation of such a sensor requires special treatment, due to the fact that the GPU is rendering the whole image at once and the number of frames per second (FPS) is limited commonly to well below 100. The solution implemented in Stonefish is based on constant FPS computations, using a shader that performs an iterative generation of all events that have happened during the time between frames. 
Its output is thus a list of all events that the EBC captured, with their respective pixel location, timestamp and polarity. The computations are fully parallelised per pixel but the list of events has to be kept consistent between pixels, which is achieved through atomics. The only CPU-based part of the implementation is sorting of the resulting events by timestamp. An example of the output is presented in Fig.~\ref{fig:evB}, where a forward-looking device is installed on a moving AUV.

\subsection{Thermal Camera}
Apart from cameras working in the visible light spectrum, there is also a common demand for the use of infrared thermal vision in the monitoring of industrial installations, as well as, surveillance. Thermal cameras are not good for underwater use but they are commonly found in drones, ground robots and surface vehicles. Therefore, a new sensor simulating the operation of such types of cameras was developed and integrated deeply into the rendering pipeline. 

Accurate simulation of heat transfer and temperature changes of bodies would require full finite element analysis (FEA), which is not feasible in a real-time framework, thus an approximation, that assumes the temperatures are measured at a steady state, was used. Moreover, a full simulation of thermal effects would require conductivity, convection and radiation calculations that would have to be done and continuously updated for each point on the surface of each body. 

\begin{figure}[t]
  \centering
  \includegraphics[width=\linewidth]{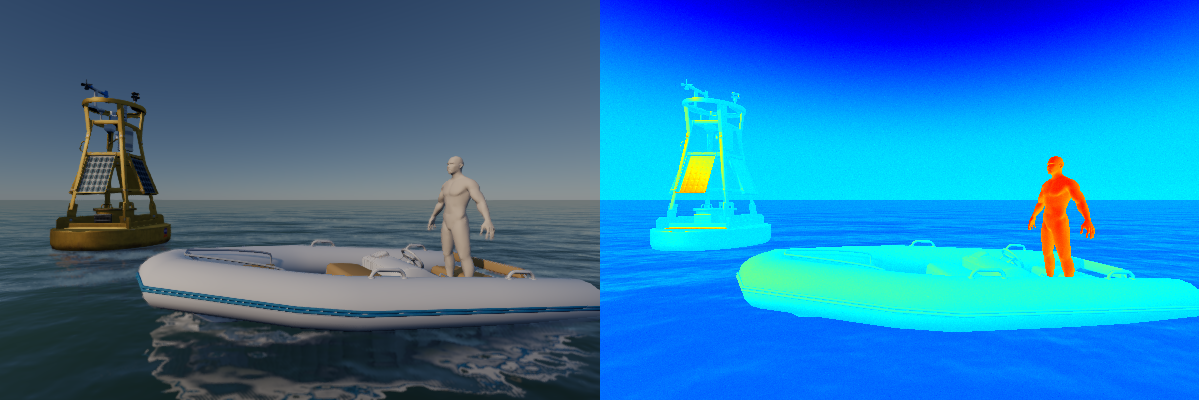}
  \caption{On the left a view from a colour camera; On the right the corresponding colour-mapped thermal image.} 
  \label{fig:thermal}
\end{figure}

The approach used in Stonefish simplifies the problem and produces a thermal image in the screen-space, not the object space. The base temperature of the body can be defined in three ways: equal to the air temperature, constant for the whole surface, or using a temperature map (texture).
The albedo and roughness of the object is determining how much of the sunlight energy is being absorbed and converted into a temperature increase. Moreover, as the rendering engine is also drawing the ocean and the sky, both of components had to be updated as well. The sky temperature is based on \cite{AWANOU1998227} and the ocean surface temperature combines water temperature, solar irradiance, and reflection of the sky. All of the temperature computations are implemented as OpenGL shaders for real-time performance and a color-mapped result can be observed in Fig.~\ref{fig:thermal}.

The user can define resolution, field of view, range of temperatures measured by the sensor, as well as, the standard deviation of the temperature measurement, which is used to include additive Gaussian noise. Moreover, the sensor produces two images: one floating-point image of temperature readings and one color-mapped image for direct display.

\subsection{Optical Flow}
\begin{figure}[t]
	\centering
    \includegraphics[width=\linewidth]{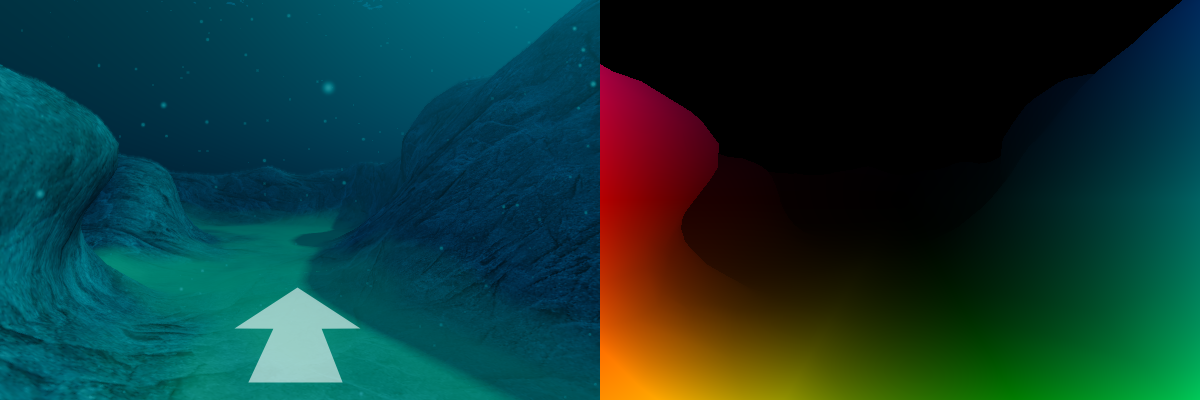}

    \caption{On the left, view from a colour camera (arrow symbolises direction of camera movement); On the right the corresponding optical flow field (colour-mapped).} 
\label{fig:oflow}
\end{figure}

Optical flow enhances navigation capabilities, facilitating more precise control and path planning for marine robots, by calculating the velocity of the objects relative to the robot's camera, 
In Stonefish, we use a shader to compute optical flow, which captures the apparent motion of objects in an image due to the relative movement between the camera and the scene. This shader integrates various camera and scene parameters to calculate the velocity of each fragment in the camera frame and subsequently projects this velocity onto the image plane.
The shader computes the velocity of each fragment by considering contributions from both the body and the camera motions. This involves determining the positions of the fragment relative to the body's centre of rotation and the camera's position and calculating the velocities induced by their respective linear and angular motions.
Subsequently, the shader projects the computed velocity onto the image plane. It determines the depth of the fragment along the view vector and computes the pixel position relative to the image centre. Utilizing the focal length, it computes the optical flow in both the X and Y directions, reflecting the apparent movement of the fragment within the image. An example of operation of the optical flow sensor is presented in Fig.~\ref{fig:oflow}.

\section{Communication devices}
Stonefish integrates several advanced existing communication technologies to allow users to test ensure reliability and adaptability of their systems.

\subsection{Acoustic Communication}
In terms of acoustic communication, it incorporates Ultra-Short Baseline (USBL) systems and acoustic modems, providing precise underwater positioning and communication capabilities over extended distances.
USBL systems facilitate precise underwater positioning by emitting acoustic signals from a vessel to an underwater transponder, which then calculates its position relative to the vessel. Acoustic modems and USBLs are simulated in Stonefish by propagating messages through the simulated environment at a speed of sound in water. This propagation follows the theoretical spherical acoustic wavefront that is created when the transducers are vibrating. Moreover, the simulator can check for straight line occlusions and take into account the transducer's field of view.

\subsection{Visual Light Communication}
\begin{figure}[t]
    \centering
    \includegraphics[width=0.48\linewidth]{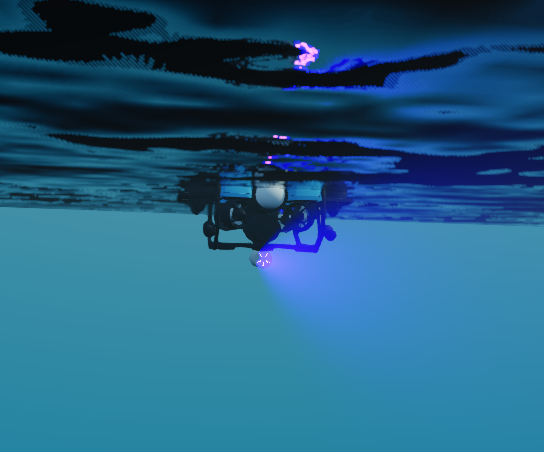}
    \includegraphics[width=0.48\linewidth]{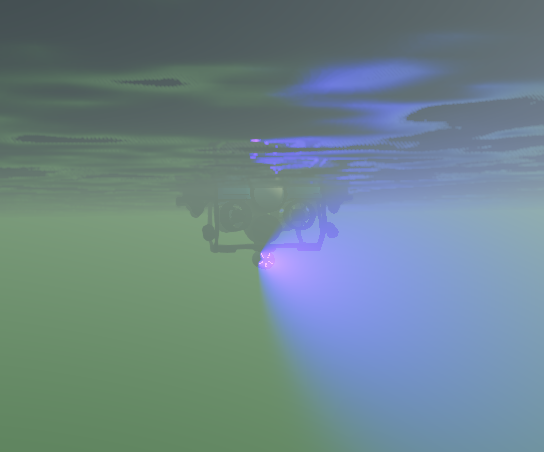}
    \caption{BlueRov2 simulation with VLC modems in clean and murky water.}
\label{fig:vlc}
\end{figure}

To go further, we also implemented optical modem communication. More specifically we developed a visual light communication (VLC) modem (see Fig.~\ref{fig:vlc}), to simulate scenarios where the robot is fully autonomous and can just communicate with the user when its VLC modem is facing another VLC modem linked with the network of a remote user. This setup is particularly useful to test "shared-autonomy" scenarios. Additionally, to simulate degraded communication conditions, we utilize ROS 2 Quality of Service (QoS), which allows us to control and manage the reliability, durability, and bandwidth usage of communication channels between robot nodes and remote users.

\section{Other significant improvements}

\subsection{Advanced Thruster Models}
A thruster actuator is an underwater device that generates thrust by moving the liquid mass, commonly using a rotating propeller. As the most prevalent actuator for underwater and surface vehicles, it requires special attention. The implemented mathematical model is modular, integrating two primary components: the rotor dynamics model and the thrust (and torque) generation model. This modularity allows for various combinations, providing a flexible setup to meet diverse user requirements. 
Concerning the rotor dynamics, we have the following models:
\begin{itemize}
    \item Zero Order: A simple passthrough with no dynamics; the input is angular velocity (rad/s).
    \item First Order: A first-order system characterized by a time constant, with angular velocity (rad/s) as input.
    \item Yoerger’s Model~\cite{yoerger1990influence}: Uses motor torque (Nm) as input, defined by parameters alpha and beta.
    \item Bessa’s Model~\cite{bessa2005thruster}: Inputs voltage (V) and includes parameters such as rotor inertia, linear and quadratic thruster constants, torque constant, and motor resistance.
    \item Mechanical PI: A mechanical model of a rotating propeller controlled using a PI controller, with angular velocity (rad/s) as input and parameters for inertia, proportional gain, integral gain, and integral limit.
\end{itemize}
For the thrust generation model, we have:
\begin{itemize}
    \item Quadratic: Utilizes a symmetrical thrust coefficient for thrust calculation.
    \item Deadband: Features asymmetrical thrust coefficients for forward and reverse, with defined deadband limits.
    \item Linear Interpolation: Transforms angular velocity into thrust based on linearly interpolated tabulated data.
    \item Fluid Dynamics: Based on advanced fluid dynamics equations, considering incoming fluid velocity, with asymmetrical thrust coefficients and an induced torque coefficient.
\end{itemize}
Implementing these diverse models allowed for accurate simulation of marine vehicle's motion. Each model caters to different operational scenarios and user requirements, from simple pass-throughs to complex fluid dynamics. This flexibility leaves to the user the decision about the underlying mathematical formulations and ensures that the simulator can be adapted to testing different control strategies.
\vspace{-0.1cm}
\subsection{Tether Cable}
Stonefish enables users to simulate surface vehicles too \cite{grimaldi2024integratingdigitaltwinconcept}. One of the scenarios including both underwater and surface vehicles is the one which includes a tether to link them. For this purpose, we simulated a cable approximated with $N$ spheres. The tether can be parameterized by defining the mass per sphere, the radius, the distance between the sphere, the length, and the damping force applied to each revolute joint, which links the spheres to each other. 
\vspace{-0.1cm}
\section{Tools and Related Developments}
\subsection{Automatic Data Annotation}
\begin{figure}[t]
    \centering
    \includegraphics[height=3.6cm,  trim=3.5cm 2.5cm 2.5cm 3.5cm,clip]{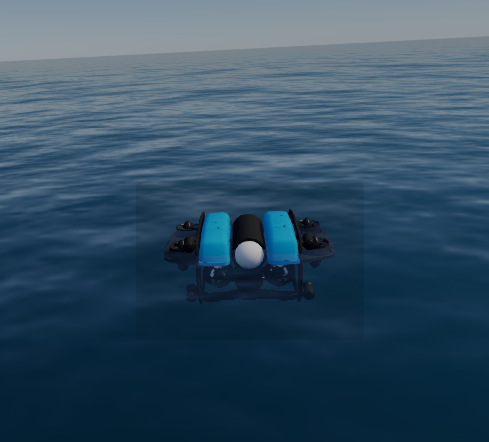}
    \includegraphics[height=3.6cm, trim=2.5cm 2.5cm 2.5cm 2.5cm,clip]{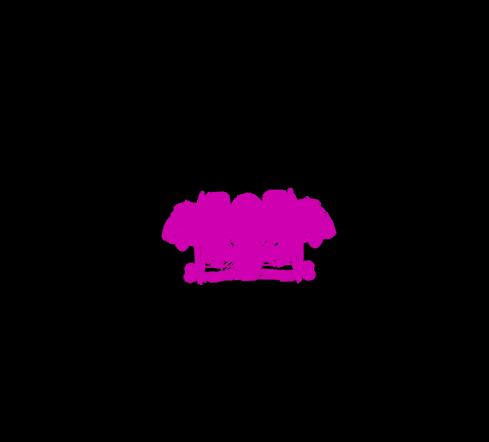}
    \caption{On the left: bounding box generated for the robot in the scene; On the right: semantic segmentation of the robot.}
    \label{fig:segment}
\end{figure}

Given the challenges of acquiring data in marine robotics—where testing at sea must be meticulously planned and can be prohibitively expensive—automatic annotation tools provide a cost-effective alternative. By automating the annotation process, these tools ensure that datasets used for training are meticulously labelled, thereby enhancing the accuracy and performance of autonomous marine robotics in practical applications.
For these reasons, we have introduced in Stonefish automatic semantic, instance and panoptic segmentation, point cloud segmentation, as well as, object detection.
Utilizing the mesh and camera view orientation, we compute the bounding boxes of the visible meshes and we annotate them using Yolo5 standard. For the point cloud segmentation, we decompose the visible meshes into a point cloud and annotate it accordingly. Concerning the semantic segmentation, using the camera view and the mesh data, we classify each pixel into meaningful categories, see Fig.~\ref{fig:segment}. Then, similarly to semantic segmentation, we identify and delineate individual object instances within the camera view and mesh data. Finally, combining both semantic and instance segmentation, we provide a comprehensive understanding of the environment by categorizing and segmenting all visible elements. 
These automated tools are designed not only to advance our own research and development but also to support the broader community by facilitating easier access to labelled data.

\subsection{OpenGym Interface}
In the underwater robotics domain, reinforcement learning's adaptability makes it particularly effective for supporting autonomous navigation and exploration, position tracking, underwater object manipulation \cite{9389378}, and adaptive control in dynamic environments. In \cite{urobench}, the authors integrated OpenAI Gym \cite{brockman2016openai} with Stonefish, using a ROS interface to connect them. However, communication through ROS has been proven to slow down the training time. To overcome this issue, we exposed the data from Stonefish to be directly accessible from OpenAI Gym using Python bindings.
Furthermore, Stonefish has a console mode which can be used in parallel running multiple instances. The console mode does not provide any functionality that requires graphics, which includes not only visualisation of the simulated scenario but also simulation of cameras, lights, depth map-based sensors and waves but provides the physics and can be used to train agents, similar to what Nvidia ISAAC GYM \cite{liang2018gpu} provides.

\section{Conclusions}
Recent advancements in the Stonefish simulator have significantly improved its effectiveness as a tool for marine robotics research and training, enhancing its capability for developing and testing ROVs and AUVs. The introduction of new GPU-accelerated sensors, improved thruster modelling, and enhanced sonar accuracy have increased the simulator's realism and functionality. The new sensor suite is also an important contribution for researchers requiring datasets for developing and testing machine learning algorithms.

Looking forward, several key developments could further expand Stonefish's utility. Firstly, transitioning from OpenGL to Vulkan for the rendering pipeline is a possible big step, as Vulkan provides means of parallel rendering to multiple contexts (e.g., multiple vision sensors), does not require creation of a window, and gives access to new functionalities of the current GPUs, like the hardware ray-tracing. Secondly, improvements in simulating the sea state and water currents, and adding weather conditions, are next important steps. Incorporating particle systems for the simulation of chemical plumes and hydrothermal vents would further broaden the Stonefish's applicability. Finally, the flexibility of the platform could further be extended by implementing a plugin interface.

\newpage
\section*{ACKNOWLEDGEMENT}
The authors gratefully acknowledge the financial support provided by Fugro for part of this research.
This work has been partially supported by the EPSRC project UNderwater IntervenTion for offshore renewable Energies (UNITE) grant number EP/X024806/1, and the project ”IURBI - Intelligent Underwater Robot for Blue Carbon Inventorying” (Ref. CNS2023-144688), funded by the Spanish Ministerio de Ciencia, Innovación y Universidades.

\bibliographystyle{ieeetr}
\bibliography{refs}

\end{document}